\documentclass[runningheads]{llncs}

\usepackage[sectionbib]{bibunits}

\usepackage{eccv}

\usepackage{eccvabbrv}

\usepackage{graphicx}
\usepackage{booktabs}

\usepackage[accsupp]{axessibility}  

\usepackage{hyperref}

\usepackage{orcidlink}

\usepackage{multirow}
\usepackage{algorithm}
\usepackage[noend]{algpseudocode}
\usepackage{placeins}

\DeclareMathOperator*{\argmax}{\arg\,\max}

\defaultbibliographystyle{splncs04} 
\defaultbibliography{main}

\newcommand{\badmetric}[1]{\textcolor{gray}{#1}}
\newcommand{\mR}[1]{\emph{mR@#1}}
\newcommand{\mNgR}[1]{\emph{mNgR@#1}}
\definecolor{MyGreen}{HTML}{66c2a5}
\definecolor{MyOrange}{HTML}{d95f02}
\definecolor{MyBlue}{HTML}{8da0cb}

\begin{document}

\title{A Fair Ranking and New Model for Panoptic Scene Graph Generation} 


\author{Julian Lorenz\orcidlink{0000-0002-8893-4355} \and
Alexander Pest \and
Daniel Kienzle\orcidlink{0000-0001-7829-1256} \and
Katja Ludwig\orcidlink{0000-0002-5721-243X} \and
Rainer Lienhart\orcidlink{0000-0003-4007-6889}}

\authorrunning{J.~Lorenz et al.}

\institute{University of Augsburg\\Augsburg, Germany\\
\email{\{julian.lorenz,alexander.pest,daniel.kienzle,\\katja.ludwig,rainer.lienhart\}@uni-a.de}\\
}

\maketitle

\begin{abstract}
  In panoptic scene graph generation (PSGG), models retrieve interactions between objects in an image which are grounded by panoptic segmentation masks.
  Previous evaluations on panoptic scene graphs have been subject to an erroneous evaluation protocol where multiple masks for the same object can lead to multiple relation distributions per mask-mask pair. This can be exploited to increase the final score. We correct this flaw and provide a fair ranking over a wide range of existing PSGG models.
  The observed scores for existing methods increase by up to 7.4 \mR{50} for all two-stage methods, while dropping by up to 19.3 \mR{50} for all one-stage methods, highlighting the importance of a correct evaluation. Contrary to recent publications, we show that existing two-stage methods are competitive to one-stage methods. Building on this, we introduce the Decoupled SceneFormer (DSFormer), a novel two-stage model that outperforms all existing scene graph models by a large margin of +11 \mR{50} and +10 \mNgR{50} on the corrected evaluation, thus setting a new SOTA. As a core design principle, DSFormer encodes subject and object masks directly into feature space.
  
\end{abstract}

\begin{bibunit}

\section{Introduction}
\begin{figure}[tb]
  \centering
  \includegraphics[width=0.9\textwidth]{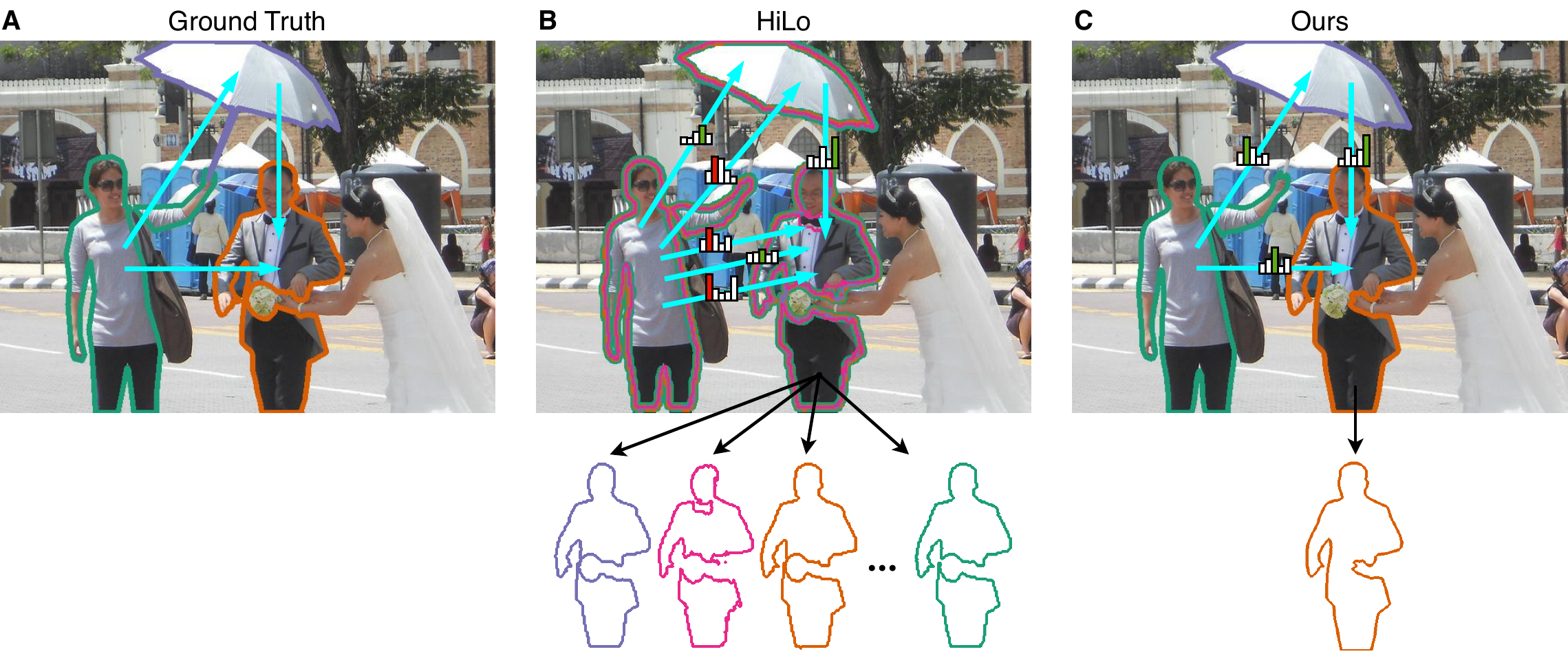}
  \caption{Schematic comparison of the output from existing one-stage methods (\eg HiLo, Fig\onedot~B) to our proposed two-stage method (Fig\onedot~C). One-stage methods often output multiple masks per real world object, visualized with colored masks in Fig\onedot~B. This results in one predicate score distribution per mask-mask pair but multiple distributions for pairs that share the same ground truth subject and object. In current evaluation implementations, multiple masks or relations are not aggregated and can therefore be exploited to increase \emph{mR@k} scores. Our new method does not have this flaw.}
  \label{fig:intro}
\end{figure}

In scene graph generation (SGG) \cite{visual_genome}, the goal is to extract a graph that represents a given image.
The nodes of the graph are the objects in the image, identified by their respective bounding box and class label. The edges of the graph are relations between the nodes which contain information about the interaction between the two nodes. A relation usually has a single predicate class assigned which has to be classified by a scene graph model.
Panoptic scene graph generation (PSGG) \cite{psg} is an extension to SGG and replaces the bounding boxes with panoptic segmentation masks\footnote{Panoptic segmentation classifies and segments every pixel in an image into semantic categories and instance identities. For more information, please refer to \cite{panseg}}.
A panoptic scene graph model extracts the segmentation masks, identifies relations between them, and assigns predicate distributions to the relations. \Cref{fig:intro}A shows an example of a panoptic scene graph.

Until now, scene graph models for PSGG have been evaluated using the definition from Yang \etal \cite{psg} which we will call \emph{Multiple Masks per Object Evaluation Protocol} (\emph{MultiMPO}). It shows two problematic peculiarities that can heavily distort the calculated scores.
First, masks of a generated scene graph are allowed to contain duplicates. In that case, a 1:1 mapping of the nodes in the graph to the real world is not possible anymore.
Second, models are allowed to output multiple predicate distributions for the same subject-object ground truth pair. A model can exploit this to increase the hit chances by predicting the same subject-object pair multiple times with different predicates, which violates the definitions of the applied metrics.
We show how to correct these two issues and name the updated and more precise evaluation protocol \emph{Single Mask Per Object Evaluation Protocol} (\emph{SingleMPO}).
Compared to the old \emph{MultiMPO}, existing PSGG one-stage models achieve much lower \mR{k} scores than previously anticipated with a decrease of up to 19.3 \mR{k}. Existing two-stage methods are invariant to the choice of the evaluation protocol and can even be boosted by using a suitable state-of-the-art segmentation model upfront.

Recent developments have shifted towards one-stage methods, \ie, inferring the graph and the masks in one pass \cite{hilo,psg,pairnet}. However, we show that with ever improving segmentation models, two-stage methods that receive the masks from a SOTA segmentation model are now able to outperform their one-stage counterparts.
To demonstrate this, we introduce the \emph{Decoupled SceneFormer} (DSFormer) that is designed from the ground up to process the outputs from a segmentation model and infer just the scene graph itself. This constraint allows us to use a much simpler network architecture that is easy to train, modify, and significantly outperforms other methods on \mR{20}, \mR{50}, \mNgR{50}, and other metrics by a large margin, setting a new SOTA in PSGG.
To summarize, we contribute the following:

\begin{enumerate}
  \item An analysis of the currently used evaluation protocol for PSGG
  \item The new Single Mask Per Object Evaluation Protocol (\emph{SingleMPO}) that corrects the flaws of the currently common evaluation protocol
  \item A thorough re-evaluation of existing methods with \emph{SingleMPO}
  \item The new Decoupled SceneFormer (DSFormer) two-stage architecture that is easy to train and substantially outperforms current state-of-the-art methods with an increase in \mR{50} of +11 points
\end{enumerate}
Code for inference, network training, and scripts to evaluate most existing panoptic scene graph models out of the box with \emph{SingleMPO} can be found here: \url{https://lorjul.github.io/fair-psgg}.

\section{Related Work}

\subsection{Datasets}
The most commonly used scene graph dataset is Visual Genome \cite{visual_genome}, but it lacks segmentation masks in its provided ground truth annotations. In this paper, we are discussing the existing flaws with the \emph{MultiMPO} evaluation protocol for methods in panoptic scene graph generation \cite{relwork_4dpsgg,psg}.
Therefore, we focus on the PSG dataset \cite{psg}. It contains 48,749 images with panoptic segmentation masks and a total of 273,618 relation annotations. Only about 3\% of all relation annotations contain multiple predicates per subject-object pair. Hence, scene graphs are usually evaluated using single-label metrics like \emph{Mean Recall@k} \cite{psg,hilo,pairnet}. In the real world, relations often have multiple predicates, \eg, a person can be both holding and drinking a bottle of water. However, panoptic scene graph generation is currently dependent on a single dataset which slows down development in the direction of efficient multi-label training.
Recently, the Haystack dataset \cite{haystack} for PSGG was proposed that tackles some of the multi-label concerns. However, the size of the dataset is only sufficient for scene graph evaluation.
Because of these limitations, metrics for PSGG have to be chosen accordingly.

\subsection{Metrics}

Originally, scene graphs were evaluated using \emph{Recall@k} (\emph{R@k}) \cite{first_scenegraph}: a model selects the top $k$ important relations in an image and assigns a single predicate class to each relation. \emph{R@k} is calculated as the number of predicates that are correct among the $k$ selected ones, divided by the total number of predicates in the image.
However, scene graph datasets have highly imbalanced predicate class frequencies, which is an active field of research \cite{relwork_invarlearn,relwork_multiproto,relwork_semproto,hilo,ietrans,relwork_cktrcm}. In this case, \emph{Mean Recall@k} (\mR{k}) is a more suitable metric \cite{kern}. It is an extension of \emph{R@k} and is calculated by collecting \emph{R@k} values per predicate per image and then averaging them at the end. This ensures that frequent predicate classes don't dominate the final score.
An extension of \emph{R@k} is \emph{No Graph Constraint Recall@k} (\emph{ngR@k}) \cite{motifs}. Instead of only allowing a single predicate class per relation, a model can distribute $k$ predicates over all available relations in an image. Multiple predicates per relation are allowed. Again, the number of correctly assigned predicates is divided by the number of predicates in the ground truth. Analogous to \mR{k}, we use \mNgR{k} which is calculated by averaging \emph{ngR@k} over all predicates. The \mNgR{k} metric can better measure multi-label ground truth and predictions because it allows multiple predicates for the same relation.
Recent SOTA methods output multiple relations for the same subject-object pairs. Despite using a single predicate per relation, they effectively assign multiple predicates per subject-object pair, confusing the \mR{k} metric with something in between \mR{k} and \mNgR{k}. With our updated evaluation protocol, this confusion is eliminated.

\subsection{Existing Methods}

In general, scene graph models can be divided into one-stage and two-stage methods.
For two-stage methods, subject and object region are given (as a box or as a mask) and the predicate has to be identified. If supported, two-stage methods can also receive the class of subject and object. When publishing the PSG dataset, Yang \etal ported four different two-stage architectures to PSGG: IMP \cite{imp}, MOTIFS \cite{motifs}, GPS-Net \cite{gpsnet}, and VCTree \cite{vctree}. We will compare our two-stage approach to these methods.
For one-stage methods, a model infers masks for subjects and objects and predicts the predicate for the extracted pairs. The first one-stage methods that were introduced for PSGG are PSGTR \cite{psg} and PSGFormer \cite{psg} which are based on the DETR \cite{detr} architecture and its extension \cite{hotr} to the Human-Object Interaction task.
Pair-Net \cite{pairnet} is another one-stage method that improves PSGFormer by splitting the graph generation process into pair detection, followed by relation classification.
The HiLo \cite{hilo} model tackles the predicate imbalance by efficiently combining rare and common predicate classes during training.
Although one-stage methods have been presented with SOTA performance, we will show that there are some caveats when evaluating them. If duplicate masks and relations are prevented, lower evaluation scores are achieved which are surpassed by two-stage models.

Two-stage methods rely on a good performance of the segmentation model in the first stage. For PSGG, we are naturally interested in panoptic segmentation models and will discuss the capabilities of Mask2Former \cite{mask2former}, MaskDINO \cite{maskdino}, and OneFormer \cite{oneformer}.
As shown in \cref{sec:firststage}, choosing a good segmentation model is crucial for two-stage methods and can almost double the performance. Two-stage methods can easily leverage foundation models for image segmentation that have been trained on datasets much larger than the available scene graph datasets.
When comparing results with existing two-stage architectures, it is therefore important to use an up-to-date segmentation model for a fair comparison.

\section{Methods}

Following the definition of Yang \etal \cite{psg}, a panoptic scene graph consists of nodes and edges. A node is a segmentation mask with a class label. An edge is a relation between two nodes that usually has a single predicate assigned. Typically, scene graph models predict a predicate class distribution for each relation and select the predicate with the highest confidence when computing a performance metric.
We now derive two essential requirements for a correct evaluation of panoptic scene graphs and analyze how they are violated by recent models. We will refer to the currently used evaluation protocol as \emph{Multiple Masks Per Object Evaluation Protocol} (\emph{MultiMPO}) and to our updated protocol as \emph{Single Mask Per Object Evaluation Protocol} (\emph{SingleMPO}).

\begin{figure}[tb]
  \centering
  \includegraphics[width=\textwidth]{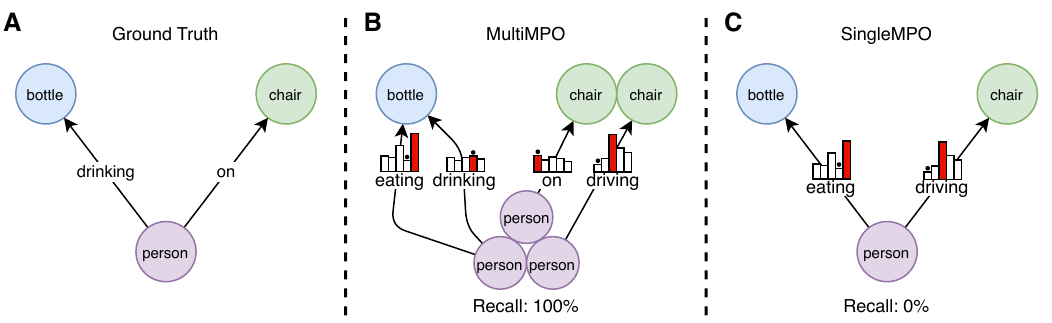}
  \caption{
    Schematic comparison of the two considered evaluation protocols.
    (\textbf{A}) The ground truth has a single mask per subject/object.
    (\textbf{B}) There are three different masks for "person" and two for "chair". Keeping them, all ground truth is covered and a recall of 100\% is computed by \emph{MultiMPO}, even though the hypothetical model in this example is much more confident with returning \emph{person-eating-bottle} instead of \emph{person-drinking-bottle} and \emph{person-driving-chair} instead of \emph{person-on-chair}
    (\textbf{C}) Enforcing a single mask per subject/object and a single predicate distribution per subject-object pair reveals the error in predicting the most probable relation.
  }
  \label{fig:overlap}
\end{figure}

\subsection{Requirements for a Fair Evaluation}
\label{sec:constraints}

\textbf{Nodes must be unique.} We argue that the goal of a potent scene graph model should be to output a connected graph of nodes. \cref{fig:overlap}B shows that this is not possible with duplicated nodes which arise if a scene graph model outputs multiple masks for the same real world entity. Hence, this must not be allowed.
However, \emph{MultiMPO} allows multiple masks to be considered as correct even if they are almost duplicates, as long as the IoU with the ground truth is larger than 50\% for each mask.
\emph{SingleMPO} on the other hand, only allows a single mask per ground truth subject/object. To merge multiple output masks from a scene graph model, we use a non-maximum suppression-like approach. Given the set of all output masks, we merge them into a set of new masks that do not overlap. At the same time, we keep track of the merge process to reassign relations to their new masks. As a result, some relations will share the same subject-object combination.

\textbf{Relations must be unique.} \Cref{fig:overlap}B shows a model that outputs two predicate distributions per subject-object pair.
Even though the model is much more confident with person-eating-bottle than with person-drinking-bottle, it will score a perfect recall with \emph{MultiMPO} because it uses both distributions, which would be incorrect with \mR{k}. Nevertheless, recent scene graph models report this score as \mR{k}, giving them an unfair advantage over models that adhere to the single predicate constraint.
With \emph{SingleMPO}, all methods are evaluated equally and the recall for the example in \cref{fig:overlap} is correctly calculated as 0\%.
If multiple predicates are indeed intended, the \mNgR{k} metric has to be used instead. Nevertheless, even for \mNgR{k}, only a single score per predicate per subject-object pair is allowed and outputting multiple distributions for the same pair would again distort the calculated metric.
To aggregate duplicate relations for a specific subject-object pair, \emph{SingleMPO} uses the highest confidence score per predicate and averages the \emph{no-relation} output.

These two postulated requirements ensure that there is no ambiguity when mapping the nodes or relations of a generated scene graph to the real world. In addition, scene graph models cannot gain unfair advantage by outputting multiple predicate distributions per subject-object pair. In contrast to \emph{MultiMPO}, \emph{SingleMPO} ensures that the requirements are always fulfilled.

\subsection{Model Overview}

Based on the steady improvements on panoptic segmentation methods \cite{mask2former,maskdino,oneformer}, we decide to use a completely decoupled two-stage approach. The first-stage model is an established segmentation model that outputs segmentation masks and class labels for a given image. Masks and labels are then used as additional inputs to our new Decoupled SceneFormer (DSFormer) architecture. Note that we only have to train the DSFormer scene graph model and not the segmentation model. This approach has four main advantages:
\begin{enumerate}
  \item Because the segmentation masks have already been inferred, a smaller model can be used to construct the scene graph, lowering hardware requirements and computational cost during training.
  \item Two-stage methods directly leverage SOTA foundation models for image segmentation without having to include them in the training pipeline. These models are trained on datasets that are much larger \cite{coco,objects365,as1b} than the available scene graph datasets and will naturally generate superior masks than one-stage scene graph models.
  \item Switching to a new segmentation model requires virtually no extra work and no retraining. When comparing new scene graph methods with existing two-stage methods, it is important to use state-of-the-art segmentation models for a fair comparison.
  \item Being able to just show selected subject-object pairs to our model during training gives us much more control over sampling strategy and loss weighting.
\end{enumerate}
\begin{figure}[tb]
  \centering
  \includegraphics[width=1.0\textwidth]{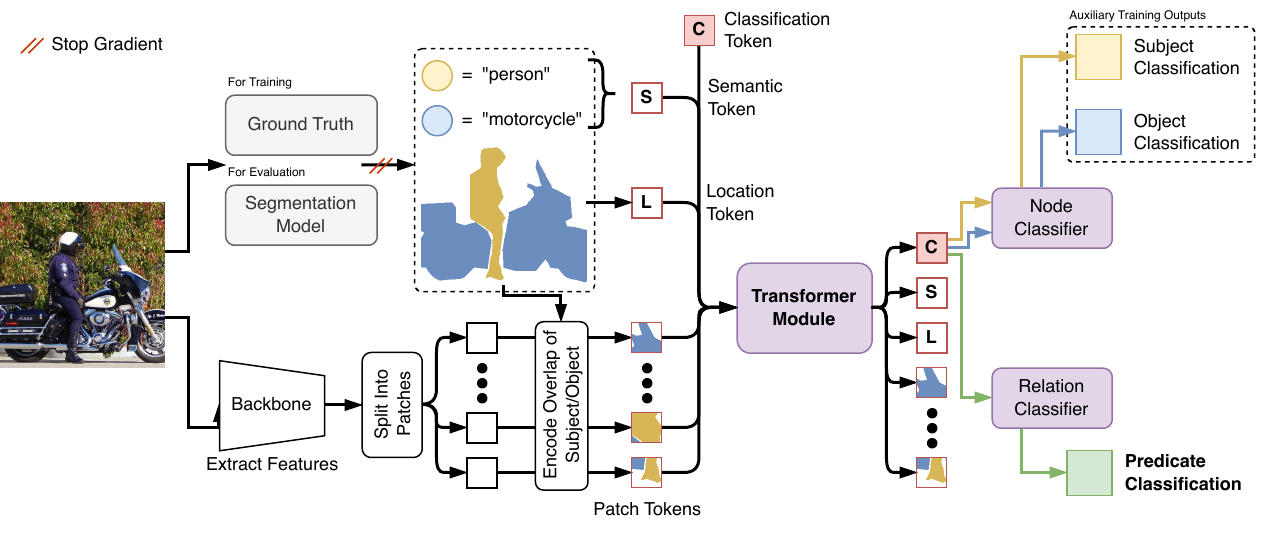}

  \caption{
    Our proposed architecture for DSFormer.
    In a forward pass, the model requires an image, subject and object class, and segmentation masks for subject and object. During training, ground truth data is used. During evaluation, segmentation masks and class labels are inferred from a capable segmentation model.
    DSFormer outputs a relation prediction as well as an auxiliary subject and object class prediction which are only used during training.
    \Cref{fig:arch_tokens} shows how the different tokens that enter the transformer module are derived.
  }
  \label{fig:arch_overview}
\end{figure}

An overview of DSFormer's architecture is depicted in \cref{fig:arch_overview}. The model is trained with ground truth segmentation masks and relations. During evaluation, we replace the ground truth masks with inferred masks from a segmentation model that is decoupled from our model.
Because DSFormer doesn't have to construct segmentation masks, its backbone is kept small and we use a ResNet-50 backbone from Faster R-CNN \cite{fasterrcnn} pretrained on object detection. To extract features, we use a feature pyramid network \cite{fpn} that outputs four different feature tensors with different resolutions, upscale them all to the largest resolution and merge them to one single feature tensor. For an RGB input image of resolution $640 \times 640$, the resulting feature tensor has a shape of $160 \times 160 \times 256$.
We split the tensor into non-overlapping patches with a patch size of $8 \times 8$ each. Each patch is projected to a token with an embedding dimension of 384. However, before all tokens can be processed by the transformer module, the location of subject and object have to be encoded.

\begin{figure}[tb]
  \centering
  \hspace{0.5cm}
  \includegraphics[width=0.7\textwidth]{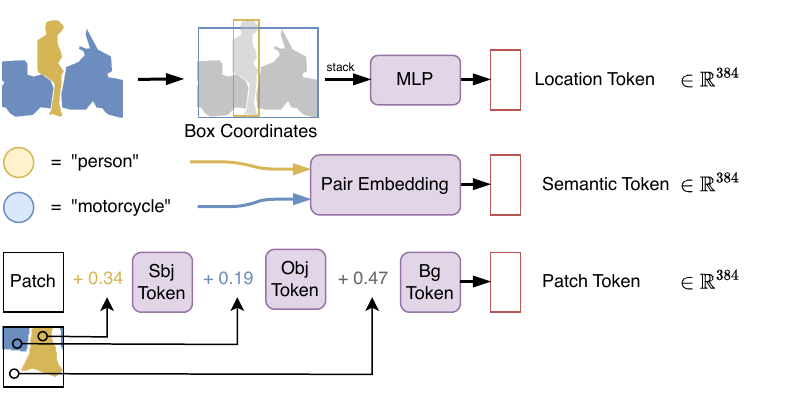}
  \caption{
    Most tokens for our proposed model are derived from the segmentation masks.
    In a patch token, the overlapping ratio of subject and object mask are encoded by adding a weighted sum over learnable subject, object, and background tokens to the initial feature patch.
    The location token is inferred from the normalized bounding boxes of subject and object using a two-layer MLP.
    The semantic token is derived directly from subject and object class via a learnable embedding that returns a unique vector for each unique subject-object class combination.
  }
  \label{fig:arch_tokens}
\end{figure}

\subsection{Subject-Object Encoding}

DSFormer has to be prompted with specific subject and object regions for which it will then return a predicate distribution that describes the relation.
For that purpose, many two-stage methods (\eg \cite{imp,gpsnet,vctree}) utilize the retrieved bounding boxes from the first-stage model and crop the feature tensor using \emph{RoIAlign} \cite{maskrcnn} or similar methods. 
Instead of using feature crops to tell DSFormer where the subject and object are located, we keep all the information from the backbone and add a prompt encoding to the patch tokens from the backbone. Therefore, DSFormer can still use global context information that lies outside of the subject or object region for the final decision. For example, an image of a restaurant is more likely to have the predicates "eating" or "drinking".

\Cref{fig:arch_tokens} visualizes how the prompt encoding is computed. It is a weighted sum of three learnable tokens $t_{sbj}$, $t_{obj}$, and $t_{bg} \in \mathbb{R}^{384}$ that encodes the presence of subject, object, and background into a specific patch token:
\begin{equation}
  \label{eq:sbjobj_enc}
  token = patch + r_{sbj} \frac{t_{sbj}}{||t_{sbj}||} + r_{obj} \frac{t_{obj}}{||t_{obj}||} + (1 - r_{sbj} - r_{obj}) \frac{t_{bg}}{||t_{bg}||}\text{ .}
\end{equation}

$r_{sbj}$ is the ratio that determines how much of the specific patch is covered by the subject mask. $r_{obj}$ is defined respectively. Subject and object mask never overlap, therefore $r_{sbj} + r_{obj} \leq 1$.
The learnable tokens are normalized by their magnitude to prevent them from dominating the patch tokens.

\subsection{Transformer Module}

At the core of DSFormer is a ViT \cite{vit} inspired transformer module with 6 layers. It receives the patch tokens and an additional learnable classification token, shown as the filled red box in \cref{fig:arch_overview}. The semantic and location tokens shown in \cref{fig:arch_overview} are optional and will be discussed in \cref{sec:addtokens}. After the transformer module, the classification token is projected to the desired relation output vector using a two-layer MLP, depicted with green arrows in \cref{fig:arch_overview}.
The output vector contains a score for each possible predicate plus one additional \emph{no-relation} class. The \emph{no-relation} predicate is a virtual predicate that does not exist in the dataset and is trained to have a high value whenever there is no annotated relation between a subject and an object.

\subsection{Relation Loss}

During training, we try to penalize false positives less than false negatives, because we cannot reliably evaluate negative ground truth from the PSG dataset.
A negative ground truth is present if (a) the predicate does not apply for the relation, (b) the annotator forgot to add the label, or (c) the annotator chose to label the relation with another predicate instead. 97\% of all relations in the PSG dataset are annotated with a single label, indicating that option (c) appears frequently.
To reduce the impact of incorrect negative ground truth, we choose a loss function that is less sensitive to false positives, shown in \cref{eq:relloss,eq:finalrelloss}.
Let $p$ be a predicate class out of all $P$ predicate classes and $n$ the sample index in a batch of size $N$. We denote $y_{n,p}$ as the ground truth, which is either 1 for positive samples or 0 for negative ones. $x_{n,p}$ is the $n$-th model output for $p$.
Then, $l_{n,p}$ is our used weighted binary cross entropy loss.
We weight the positive samples with $w_p = \frac{\text{\# all neg training samples for } p}{\text{\# all pos training samples for } p}$.
As a consequence, incorrect negative ground truth has a very low impact on the training loss but the model still learns to differentiate between positive and negative because of the abundance of negative ground truth.

\begin{align}
  \label{eq:relloss}
  l_{n,p} & = - \left(w_p y_{n,p} \cdot log\ \sigma(x_{n,p}) + (1-y_{n,p}) \cdot log(1-\sigma(x_{n,p}))\right) \\
  \label{eq:finalrelloss}
  \mathcal{L}_{rel} & = \frac{1}{N \cdot P} \sum_{n=1}^{N} \sum_{p=1}^{P} l_{n,p}
\end{align}

To provide positive samples for the \emph{no-relation} predicate, we sample additional subject-object pairs without an annotation in the dataset. Each pair is then labelled with a positive \emph{no-relation} ground truth, while the remaining predicates are labelled negative. These pairs are then included in the training set to evenly balance the number of positive and negative \emph{no-relation} samples.

\subsection{Auxiliary Node Loss}
\label{sec:nodeloss}

In addition to the relation loss, we employ an auxiliary node loss that helps the model with better understanding the prompted subject and object regions.
Therefore, DSFormer outputs a subject and object classification output that is projected from the transformer module's classification token by a shared Node Classifier, which is a two-layer MLP. The process is shown in \cref{fig:arch_overview} with yellow and blue lines respectively. For the two outputs, we use cross-entropy loss, weighted with the inverse frequency of each class. The two losses $\mathcal{L}_{sbj}$ and $\mathcal{L}_{obj}$ are then averaged and return the node loss as $\mathcal{L}_{node} = \frac{\mathcal{L}_{sbj} + \mathcal{L}_{obj}}{2}$.
The final training loss is a weighted sum of $\mathcal{L} = \lambda_{rel} \cdot \mathcal{L}_{rel} + \lambda_{node} \cdot \mathcal{L}_{node}$ and we choose $\lambda_{rel} = 0.8$ and $\lambda_{node} = 0.2$.

\subsection{Additional Input Tokens}
\label{sec:addtokens}

Some relations like \emph{on}, \emph{attached to}, \emph{holding} can benefit from additional information about the location of subject and object. Therefore, we derive an additional location token as shown in \cref{fig:arch_tokens} in every forward pass, which is inspired by \cite{arbitrary_keypoints}.
We use the inferred subject and object masks of the first-stage model or ground truth to calculate the respective bounding boxes and normalize them to the range of $[-1, +1]$. We concatenate all box coordinates into a single vector
$({x_1 y_1 x_2 y_2}^{(sbj)}\ {x_1 y_1 x_2 y_2}^{(obj)})^T \in \mathbb{R}^8$
and project this vector using a two-layer MLP to an additional token for the transformer module.

It has been shown that encoding information about the subject and object classes into scene graph generation increases performance on predicate classification \cite{motifs}. To encode this semantic information, DSFormer learns a unique token vector for each combination of subject class and object class. During training, the subject and object classes are provided by the ground truth and used to select the correct token. The token is then passed as an additional token to the transformer module. During inference, the subject and object classes are provided by the first-stage model.
If DSFormer is intended to work with unknown subject/object classes, the semantic token has to be removed. This slightly decreases performance but enables zero-shot prompting.

\subsection{Evaluation}

In inference, DSFormer is run on every possible subject-object pair in an image. This can be achieved in reasonable time because the output features from the backbone only have to be inferred once per image. The result is a list of all possible relations with a predicate distribution including the virtual \emph{no-relation} predicate.
For \emph{Mean Recall@k} (\mR{k}), a scene graph model has to output a list of $k$ subject-predicate-object triplets per image that have to cover as much ground truth annotation as possible.
DSFormer selects the output relations with the $k$ lowest \emph{no-relation} scores.
Next, for each selected relation, the argmax over the other predicate scores is used to determine the predicate for the triplet.
For \emph{Mean No-Graph-Constraint Recall@k} \cite{motifs} (\mNgR{k}), a scene graph model again has to output a list of $k$ subject-predicate-object triplets per image, but the same subject-object combination is allowed multiple times as long as the predicates are different. For every predicate $p$ in output relation $r$, DSFormer combines the estimated predicate score $s_{r,p}$ with the estimated \emph{no-relation} score $s_{r,no}$ of the same relation into a ranking score $x_{r,p}$:

\begin{align}
  x_{r,p} = (1 - \sigma(s_{r,no})) \cdot \sigma(s_{r,p}) && \sigma \text{ is the sigmoid function.}
\end{align}

Next, DSFormer sorts all $x_{r,p}$ scores within an image and selects the top $k$ ones. The $r$ and $p$ values are used to derive the returned subject-predicate-object triplets.

\section{Experiments}

\begin{figure}[tb]
  \centering
  \includegraphics[width=1.0\textwidth]{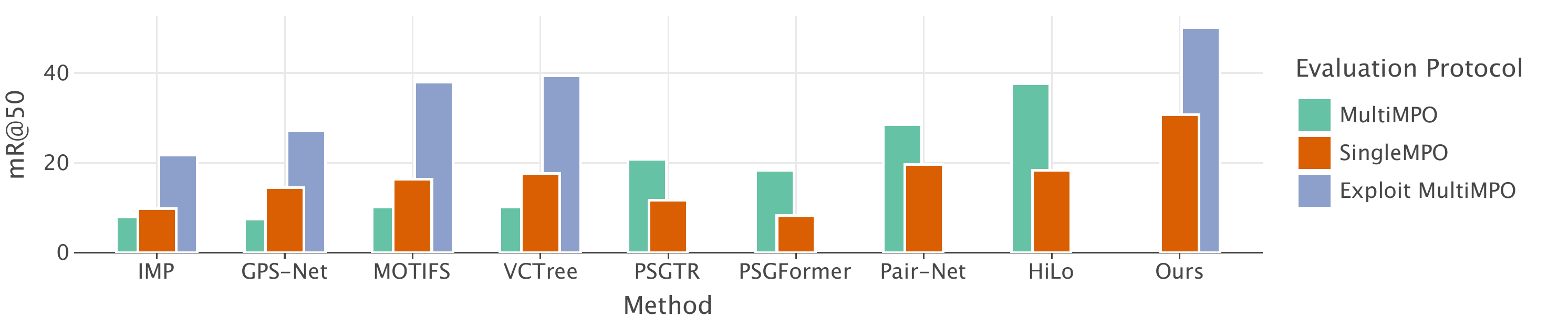}

  \caption{Comparison of achieved \mR{50} scores with: \textcolor{MyGreen}{(1)} originally published unfair \emph{MultiMPO}, \textcolor{MyOrange}{(2)} our newly introduced fair \emph{SingleMPO}, and \textcolor{MyBlue}{(3)} a modification of two-stage methods that uses a better mask model and exploits \emph{MultiMPO} similar to some one-stage methods.
  Even though all methods are evaluated equally, \mR{50} scores for all one-stage methods decline with a maximum decrease of 19.3 for \emph{SingleMPO}.
  }
  \label{fig:protocolchange}

  \vspace{-11pt}

\end{figure}

In this section, we re-evaluate the different PSGG approaches, including our novel DSFormer.
If available, we use published model weights from the authors. If not, we train the models as described in the respective publication. We then generate predictions using \emph{SingleMPO}, \ie, by merging segmentation masks that describe the same visual object and by deduplicating relations between the same subject-object pair as described in \cref{sec:constraints}. Next, we evaluate the various approaches by using \mR{k} and \mNgR{k} on the usual Predicate Classification (\emph{PredCls}) and Scene Graph Generation (\emph{SGGen}) tasks and show the results in \cref{tab:general_perf}.
In \emph{SGGen}, a model has to infer subject/object masks, class labels, and relations on its own. In \emph{PredCls}, ground truth masks and mask labels are provided and just the correct predicate classes have to be retrieved for each relation. One-stage methods cannot be prompted with given subject/object masks and can consequently not be evaluated on \emph{PredCls}.

\begin{table}[tb]
  \caption{Performance comparison on the PSG dataset\cite{psg} using \emph{SingleMPO} with \emph{Mean Recall@k} (\mR{20} and \mR{50}) and \emph{Mean No-Graph-Constraint Recall@k}\cite{motifs} (\mNgR{50}). Higher scores are better. We obtain lower values for PSGTR, PSGFormer, Pair-Net, and HiLo because multiple relation outputs for the same subject-object pair are removed. Missing values indicate that the respective model is a one-stage model and cannot be evaluated on Predicate Classification. The MultiMPO column shows the previously incorrectly calculated \mR{50} scores. Technically, our DSFormer method inherently uses \emph{SingleMPO}. However, its output can be post-processed to exploit \emph{MultiMPO} (described in the supplementary). This score is shown in parantheses.}
  \label{tab:general_perf}
  \centering
  \resizebox{0.8\linewidth}{!}{
  \begin{tabular}{lrrr@{\hskip 8pt}rrrr}
    \toprule
     & \multicolumn{3}{c@{\hskip 8pt}}{Predicate Classification $\uparrow$} & \multicolumn{4}{c}{Scene Graph Generation $\uparrow$} \\
    Method & mR@20 & mR@50 & mNgR@50 & mR@20 & mR@50 & mNgR@50 & \badmetric{MultiMPO} \\
    \midrule
    IMP & 11.25 & 12.72 & 27.58 & 8.81 & 9.78 & 21.73 & \badmetric{7.88} \\
    MOTIFS & 20.00 & 21.83 & 47.98 & 15.10 & 16.32 & 37.96 & \badmetric{10.10} \\
    GPS-Net & 15.46 & 18.62 & 33.60 & 12.35 & 14.48 & 27.14 & \badmetric{7.49} \\
    VCTree & 21.19 & 23.07 & 50.24 & 16.29 & 17.58 & 39.41 & \badmetric{10.20} \\
    PSGTR & - & - & - & 10.93 & 11.62 & 27.57 & \badmetric{20.80} \\
    PSGFormer & - & - & - & 8.20 & 8.20 & 21.75 & \badmetric{18.30} \\
    Pair-Net & - & - & - & 18.02 & 19.64 & 21.48 & \badmetric{28.50} \\
    HiLo & - & - & - & 17.51 & 18.33 & 40.48 & \badmetric{37.60} \\
    \textbf{Ours} & \textbf{34.03} & \textbf{40.06} & \textbf{64.05} & \textbf{27.20} & \textbf{30.67} & \textbf{50.08} & \textbf{\badmetric{(50.08)}} \\
    \bottomrule
    \end{tabular}
  }

    \vspace{-6pt}

\end{table}

As can be seen in \cref{tab:general_perf}, we observe about the same \emph{PredCls} scores for existing two-stage methods as reported in \cite{psg} where the less rigorous \emph{MultiMPO} was used. The consistent scores are expected because these two-stage methods do not output overlapping masks and don't exploit duplicate relation predictions. Thus, they already inherently use \emph{SingleMPO}.
On \emph{SGGen}, our reported values differ greatly from the original work.
\Cref{fig:protocolchange} shows how heavily the \mR{50} scores can be distorted, if the wrong evaluation protocol is selected.
Existing one-stage methods output multiple masks per ground truth and duplicate relations which should not be allowed for the final metric. After merging the masks, PSGTR contains on average 4.19 duplicate relations per image that are removed in \emph{SingleMPO}, PSGFormer has 89.36, Pair-Net has 23.44, and HiLo has 36.08. Aggregating these duplicates as described in \cref{sec:constraints} reveals that \mR{50} scores for all one-stage models decline with a maximum decrease of 19.3 \mR{50} lower than previously reported.
Existing two-stage methods on the other hand are not affected because they already adhere to \emph{SingleMPO} as discussed before. For a fairer comparison with SOTA models, we replace the first-stage model for every two-stage method with the top-performing MaskDINO segmentation model to obtain better segmentation masks. This increases \mR{50} of \emph{VCTree} by 7.4 and almost doubles it for \emph{GPS-Net}. We will discuss the choice of first-stage model in more detail in \cref{sec:firststage}.

\begin{figure}[tb]
  \centering
  \begin{minipage}[t]{0.45\linewidth}
    \includegraphics[width=0.8\linewidth]{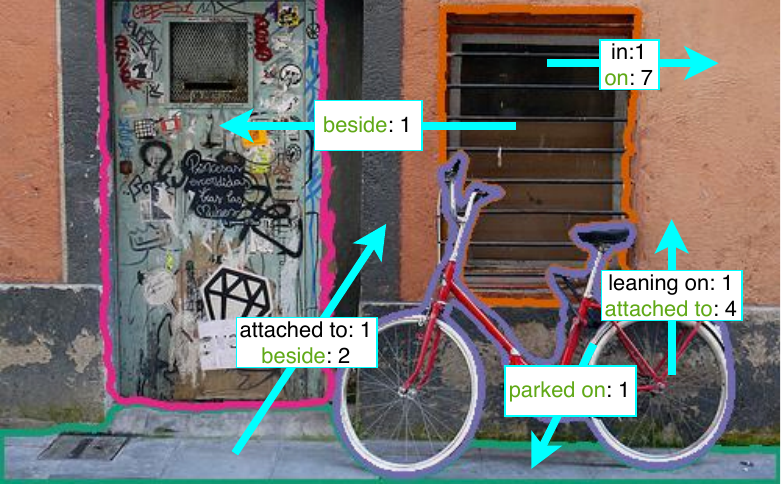}
    \centering
    \caption{Example output of DSFormer. The numbers indicate how DSFormer sorts the predicates within a relation. More images are shown in the supplementary.}
    \label{fig:output-image}
  \end{minipage}
  \hfill
  \begin{minipage}[t]{0.5\linewidth}
    \includegraphics[width=0.9\linewidth]{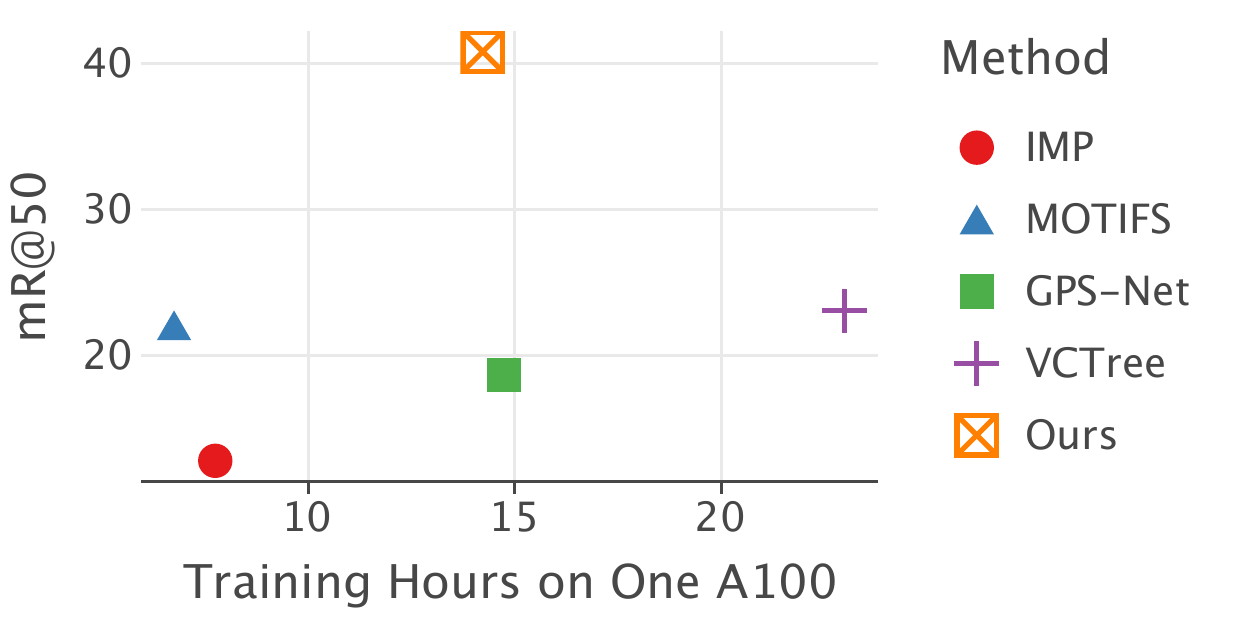}
    \centering
    \caption{Comparison of training times of existing two-stage methods on a single A100 GPU. Our method achieves a much higher \mR{50} while keeping the training time comparably low.}
    \label{fig:trainingtime}
  \end{minipage}%
\end{figure}

Contrary to recent developments, we demonstrate that two-stage methods outperform one-stage methods easily in a fair comparison. Our DSFormer model achieves SOTA performance on all reported metrics, with +11 \mR{50} and +10 \mNgR{50} compared to the previous state-of-the-art on \emph{SGGen}. Compared to one-stage models, DSFormer increases \mR{50} by more than 50\%. On top of the outstanding performance, training DSFormer is fast as can be seen in \cref{fig:trainingtime}.

\subsection{Influence of First-Stage Models}
\label{sec:firststage}

\begin{figure}[tb]
  \centering
  \includegraphics[width=0.7\textwidth]{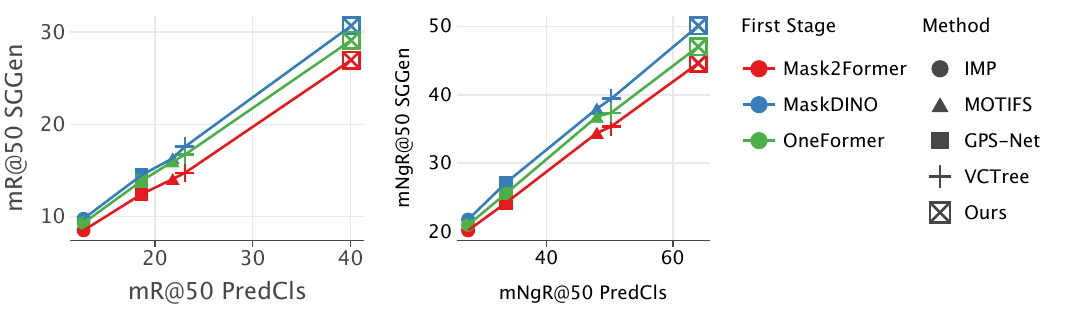}
  \caption{Performance on \emph{PredCls} (without first-stage model) is directly proportional to \emph{SGGen} (with first-stage model) apart from small fluctuations. For all tested two-stage methods, MaskDINO works best.}
  \label{fig:firststage_corr}

  \vspace{-10pt}

\end{figure}

When evaluating two-stage methods on PSGG, a good segmentation model is essential for a good overall scene graph performance. \Cref{fig:firststage_corr} shows how \mR{50} and \mNgR{50} performance on \emph{SGGen} is directly proportional to \mR{50} and \mNgR{50} performance on \emph{PredCls}, regardless of the used segmentation model. Existing two-stage methods are not specifically targeted towards certain segmentation models, which indicates that the first stage can be swapped easily.

\begin{figure}[tb]
  \centering
  \includegraphics[width=0.7\textwidth]{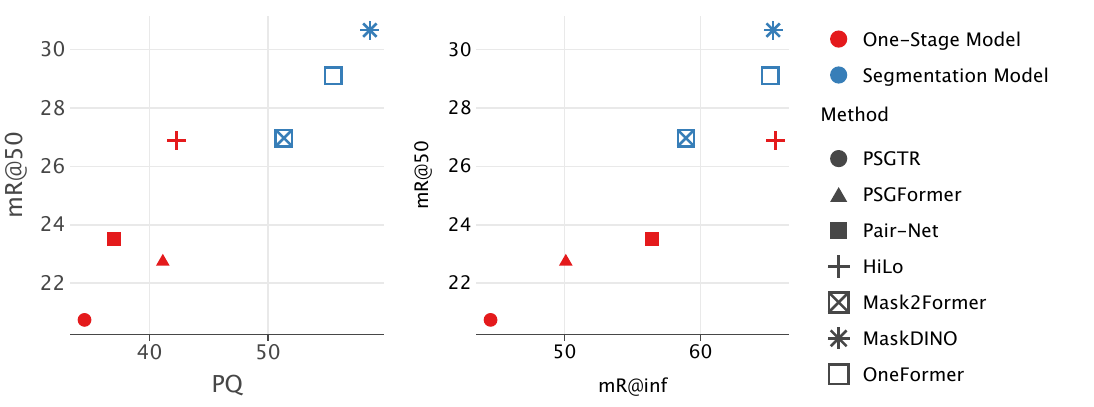}
  \caption{Performance of the used first-stage models on the PSG dataset. In addition to Panoptic Quality (PQ), we report \emph{mR@inf} as the highest possible mean recall that a subsequent scene graph model can possibly reach with the inferred masks. For comparison, we interpret the extracted masks from HiLo, Pair-Net, PSGTR, and PSGFormer as a first-stage model. The y-axis shows the best \mR{50} achieved with DSFormer as the second stage. All shown \mR{50} scores outperform previous SOTA scores.}
  \label{fig:maskmodels}
\end{figure}

In \cref{fig:maskmodels}, we compare the Mask2\-Former\cite{mask2former}, MaskDINO\cite{maskdino}, and OneFormer\cite{oneformer} segmentation models. In addition, we use the segmentation outputs of the PSGTR, PSGFormer, Pair-Net, and HiLo one-stage models. If combined with DSFormer, a segmentation model with a high Panoptic Quality (PQ)\footnote{PQ is a standard metric for panoptic segmentation that measures segmentation quality and recognition quality.} \cite{panseg} enables a better \mR{50} in general. HiLo is an outlier and generates segmentation masks with a low PQ which nevertheless help DSFormer to reach a good \mR{50} of 26.89.
To explain this behavior, we design a measure called \mR{inf}. This metric pretends that there is a perfect scene graph model after the segmentation model (or a model that has $k = \infty$ guesses). Given the extracted segmentation masks, \mR{inf} calculates what the best \mR{k} for any $k$ would be. A segmentation model with high \mR{inf} is good at retrieving masks that are relevant to improve the \mR{k} metric. Again, \cref{fig:maskmodels} shows a correlation between \mR{inf} and \mR{50} with DSFormer. In fact, HiLo achieves the best \mR{inf} compared to all other models. A perfect scene graph model would perform best on segmentation masks from HiLo.
However, we suspect that segmentation models with high \mR{inf} but low PQ would be very good in theory but they make it more difficult to correctly prompt a subsequent scene graph model. To analyze segmentation models, PQ and \mR{inf} should always be used together. The best segmentation models that enable the highest \mR{50} for DSFormer are MaskDINO with a score of 30.67, followed by OneFormer (29.10), and Mask2Former (26.97). HiLo is the only one-stage model that gets close with a score of 26.89.

\subsection{Ablation Study}
\label{sec:ablation}

The effect of our introduced components can be seen in \cref{tab:ablation}. As a baseline, we use DSFormer that receives rectangular segmentation masks, derived from the bounding boxes. Adding full mask information, as well as semantic and location information via additional tokens as described in \cref{sec:addtokens}, increases performance when applied individually or in combination.
%
\begin{table}[tb]
  \caption{Improvements caused by the additional semantic and location token. All reported values are calculated for Predicate Classification (PredCls) on the PSG dataset.
  For the first row, we replace the input segmentation masks for DSFormer with rectangular masks, derived from bounding boxes.}
  \label{tab:ablation} 
  \centering
  \resizebox{0.6\linewidth}{!}{
  \begin{tabular}{c@{\hskip 8pt}c@{\hskip 8pt}c@{\hskip 10pt}rr}
    \toprule
    Masks & Semantic Token & Location Token & mR@50 & mNgR@50 \\
    \midrule
    \textcolor{red}{$\times$} & \textcolor{red}{$\times$} & \textcolor{red}{$\times$} & 34.35 & 53.48 \\
    \textcolor{Green}{\checkmark} & \textcolor{Green}{\checkmark} & \textcolor{red}{$\times$} & 36.64 & 58.36 \\
    \textcolor{Green}{\checkmark} & \textcolor{red}{$\times$} & \textcolor{Green}{\checkmark} & 38.78 & 61.75 \\
    \textcolor{Green}{\checkmark} & \textcolor{Green}{\checkmark} & \textcolor{Green}{\checkmark} & 40.06 & 64.05 \\
    \bottomrule
    \end{tabular}
  }

    \vspace{-10pt}

\end{table}
In addition, we observe that removing the auxiliary loss from \cref{sec:nodeloss}, reduces \mR{50} from 40.06 to 34.18, highlighting the need for additional guidance on subject/object information. A more extensive ablation study can be found in the supplementary.

\section{Conclusion}

We have identified unexpected and undesirable effects with the current evaluation protocol of panoptic scene graph generation and discussed the requirements that a good evaluation protocol should fulfil. We believe that our updated \emph{SingleMPO} more accurately captures what makes a good panoptic scene graph model and suggest switching to our protocol instead.
Furthermore, we showed that if we correct the current flaws in the evaluation, existing one-stage methods achieve much lower scores than previously reported whereas two-stage methods confirm their reported scores.
We introduced DSFormer, a new two-stage architecture that is completely decoupled from the used segmentation model. It can be prompted with subject and object masks from any segmentation system. It uses a specialized patch encoding and outperforms all other scene graph models with a \mR{50} of 30.67 (+11) and a \mNgR{50} of 50.08 (+10), thus setting a new SOTA performance.
To further improve its performance, DSFormer could be pretrained or extended with external knowledge \cite{relwork_cktrcm,vlprompt,relwork_semproto}. Future work should also investigate on how to leverage information between relation pairs.

As panoptic segmentation models progress, we advocate for two-stage scene graph methods as promising candidates for future top-performing PSGG methods. Selecting an up-to-date first-stage segmentation model is crucial for fair comparisons, as \emph{SGGen} scores will improve accordingly.

\putbib

\end{bibunit}

\clearpage
\appendix

\begin{bibunit}

\section{Definition of Mean Recall@k}

Calculating the discussed scene graph metrics with \emph{SingleMPO} can be split into two steps.
First, a set of subject-predicate-object triplets is retrieved from the model output.
Second, the predicted segmentation masks are matched with the ground truth.
The set of matched output triplets is used to calculate the recall-based metrics.

\emph{mR@k} and \emph{mNgR@k} are defined per image. To calculate the score over the whole dataset, the per-image metric scores are averaged.

For a given image, we define

\begin{itemize}
  \item $M_{gt}$: Set of ground truth masks that describes the visual objects in an image.
  \item $M_{out}$: Set of predicted masks
  \item $P$: Set of all possible predicate classes in the dataset. For example, the PSG dataset contains 56 predicate classes.
  \item $G$: Set of ground truth subject-predicate-object triplets. We define such triplets as $(t_{sbj} \in M_{gt}, t_{pred} \in P, t_{obj} \in M_{gt})$.
  \item $X_k$: Set of top $k$ subject-predicate-object triplets. Triplets are defined as $(t_{sbj} \in M_{out}, t_{pred} \in P, t_{obj} \in M_{out})$. The model decides what the top $k$ triplets are. For example, DSFormer uses its \emph{no-relation} output.
\end{itemize}

\subsection{Mask Matching}
\label{sec:matching}

To calculate the metrics, we match the predicted segmentation masks to the ground truth, such that each ground truth mask has at most one predicted mask assigned to. If no predicted segmentation mask overlaps with a ground truth mask with an IoU greater than 0.5, no predicted mask is assigned to the ground truth. In the following, this mapping is called $L$. See \cref{alg:match} for an explanation of the matching process. In the process, some masks from $M_{out}$ cannot be matched. Any predicted relations that are connected to those unassigned masks are discarded.

\begin{algorithm}[H]
  \caption{Mask Matching}
  \label{alg:match}
  \begin{algorithmic}[1]
    \State \textbf{Input:} Predicted masks $M_{out}$, ground truth masks $M_{gt}$, minimum IoU threshold $t$
    \State \textbf{Output:} Lookup table $L$ that maps masks from $M_{out}$ to masks in $M_{gt}$
    \Procedure{matching}{$M_{out}, M_{gt}, t$}
      \State Initialize lookup table $J$ to $J[x] = null\ \forall x \in M_{gt}$
      \ForAll{$m$ in $M_{out}$}
        \State $x \gets \argmax_{g \in M_{gt}} iou(g, m)$
        \If {$iou(x, m) > t$}
          \If {$J[x]\ \text{is}\ null \textbf{ or } iou(x, m) > iou(x, J[x])$}
            \State $J[x] \gets m$
          \EndIf
        \EndIf
      \EndFor
      \State $L \gets J^{-1}$ \Comment Use the inverse mapping of $J$
      \State \textbf{return} $L$
    \EndProcedure
  \end{algorithmic}
\end{algorithm}


\subsection{Metric Definitions for mR@k and mNgR@k}

A scene graph model usually returns a set of predicate distributions per subject-object pair and can be sorted and converted to a set of the top $k$ important subject-predicate-object triplets $X_k$. How the most probable triplets are selected is up to the model and not part of the metrics.
Using the matching process described in \cref{sec:matching}, $X_k$ can be matched to ground truth segmentation masks. If a subject/object mask cannot be matched, the whole triplet is removed. After the matching, $X'_k$ is the matched set of triplets with the matched segmentation masks and the predicted predicate classes.

We define $G^{(p)} \subset G$ as the ground truth subset that only contains triplets with predicate $p$. The model output subset $X^{(p)}_k \subset X'_k$ contains only triplets with predicate $p$. Therefore $\bigcup_{p \in P} X^{(p)}_k = X'_k$.

For Mean Recall@k (\emph{mR@k}), the matched model output $X'_k$ \textbf{must not contain any two triplets that share the same subject and object}.
This constraint is not fulfilled with \emph{MultiMPO} and thus leads to incorrect metric scores.
\emph{mR@k} is defined as:

\begin{equation}
  mR@k = \frac{1}{|P|} \sum_{p \in P} \frac{|G^{(p)} \cap X^{(p)}_k|}{|G^{(p)}|}
\end{equation}

Mean No Graph Constraint Recall@k (\emph{mNgR@k}) is calculated like \emph{mR@k} except that in contrast to \emph{mR@k}, the matched model output $X'_k$ may contain two or more triplets that share the same subject and object as long as the predicates are different.

\Cref{alg:recall} shows how Mean Recall is calculated.

\begin{algorithm}[tb]
  \caption{Mean Recall for a Single Image}
  \label{alg:recall}
  \begin{algorithmic}[1]
    \State \textbf{Input:} Set of all predicate classes $P$, ground truth triplets $G$, \\lookup table $L{:\ } M_{out} \rightarrow M_{gt}$ (\cref{alg:match}), top $k$ predicted triplets $X_k$
    \State \textbf{Output:} Mean Recall@k
    \Procedure{MeanRecall}{$P, G, L, X_k$}
      \ForAll{$p$ in $P$}
        \State $G^{(p)} \gets \{t \in G | t_{predicate} = p\}$
        \State $X^{(p)} \gets \{\}$
      \EndFor
      \ForAll{$t$ in $X_k$} \Comment Match predicted masks to ground truth
        \If{$t_{sbj} \in L$ and $t_{obj} \in L$}
          \State $p \gets t_{predicate}$
          \State $t' \gets (L[t_{sbj}], p, L[t_{obj}])$
          \State $X^{(p)} \gets X^{(p)} \cup t'$
        \EndIf
      \EndFor
      \State \textbf{return} $\frac{1}{|P|} \sum_{p \in P} \frac{|G^{(p)} \cap X^{(p)}_k|}{|G^{(p)}|}$

      \Comment For the whole dataset, calculate Mean Recall for every image, then average
    \EndProcedure
  \end{algorithmic}
\end{algorithm}

\section{Evaluating on MultiMPO With DSFormer}

\begin{algorithm}[tb]
  \caption{Convert SingleMPO Output to MultiMPO Output}
  \label{alg:convert}
  \begin{algorithmic}[1]
  \State \textbf{Input:} List $R$ of relation outputs; set of predicate classes $P$ (excluding the no-relation class)
  \State \textbf{Output:} Modified list $R'$ for \emph{MultiMPO}
  
  \Procedure{Convert}{$R, P$}
    \State Initialize empty list $R'$
    \ForAll{$\text{relation } r \text{ in } R$}
      \ForAll{$\text{predicate } p \text{ in } P$}
        \State Initialize new relation $r'$
        \State $r'[norel] \gets (1 - (1 - r[norel]) \cdot r[p])$ \Comment Assign no-graph-constraint score
        \State $r'[p] \gets 1$ \Comment This ensures that the argmax is $p$
        \ForAll{$\text{predicate } q \text{ in } P \setminus p$}
          \State $r'[q] \gets 0$
        \EndFor
        \State Add $r'$ to $R'$
      \EndFor
    \EndFor
    \State \textbf{return} $R'$
  \EndProcedure
  \end{algorithmic}
  \end{algorithm}

DSFormer adheres to the \emph{SingleMPO} evaluation protocol by default. As we have discussed, it does not make sense to use \emph{MultiMPO} for evaluation. However, to prove that \emph{MultiMPO} can be used to gain an unfair advantage over models that already inherently adhere to \emph{SingleMPO}, we can post-process the \emph{SingleMPO} output of DSFormer and convert it to \emph{MultiMPO} as depicted in \cref{alg:convert}. For each predicate $p$ in each relation $r = (r_{norel}, r_1, \dotsc, r_P)$, we create a new relation $r' = (r'_{norel}, r'_1, \dotsc, r'_P)$ that has a score of $r'_p = 1$ for predicate $p$ and a \emph{no-relation} score of $r'_{norel} = 1 - ((1 - r_{norel}) \cdot r_p)$. DSFormer uses the \emph{no-relation} score to determine the top $k$ triplets for the recall metrics. With the new constructed \emph{no-relation} score, the top $k$ mNgR@k triplets are now the top $k$ mR@k triplets (only when using \emph{MultiMPO}).

\section{Model Architecture Details}

\subsection{Parameters}

If not specified otherwise, we used the following parameters during training:
For the transformer module, we use 6 transformer layers with an embedding dimension of 384. We add a 2D-sine positional encoding in every layer as in \cite{transformer}.
We use a batch size of 32 and train with AdamW\cite{adamw} with a learning rate of $3.7 \times 10^{-5}$ and weight decay of 0.04.
The subject/object encoding (discussed in Sec\onedot 3.3) is added once to the patch tokens before they enter the transformer module.
We resize the input images to a resolution of $640 \times 640$.

\subsection{Inference Speed}

For $n$ masks, $n^2 - n$ relations must be classified to generate a complete scene graph.
Existing one-stage methods circumvent this issue by limiting the number of relations to a fixed number (usually 100 relations per image), resulting in incomplete scene graphs. With DSFormer on the other hand, we choose to generate the complete scene graph because we expect it to be more useful for downstream tasks.
In practice, this approach is feasible as shown in \cref{tab:inf_speed}.
On a single NVIDIA A100, our implementation can process about 2400 relations in one forward pass, which is sufficiently fast.
Additional relations can be processed sequentially without computing the feature tensor again.
Only 0.2\% of all images in the PSG dataset have to be split.

\begin{table}
  \caption{Comparison of number of learnable parameters and required time to run inference on the full test set. Each method was evaluated on a single A100 GPU.}
  \label{tab:inf_speed}
  \centering
  \begin{tabular}{l r r}
    \toprule
    Method & \# Parameters & Inference \\
    \midrule
    IMP & 78M & 2.2 min \\ 
    MOTIFS & 108M & 1.3 min \\ 
    GPS-Net & 82M & 3.4 min \\ 
    VCTree & 104M & 4.4 min \\ 
    PSGTR & 44M & 4.8 min \\ 
    PSGFormer & 52M & 3.9 min \\ 
    Pair-Net & 54M & 3.6 min \\ 
    HiLo & 230M & 15.2 min \\ 
    \textbf{Ours} & 50M & 8.4 min \\  
    \bottomrule
\end{tabular}
\end{table}

\subsection{Location Token}

Given a segmentation mask for the subject and a segmentation mask for the object, we calculate two bounding boxes. We use $x_1$, $y_1$, $x_2$, $y_2$ to denote left, top, right, and bottom coordinates of the bounding box. To normalize them, we divide by the image width $w$ or height $h$ and normalize the coordinates to the range of $[-1, +1]$:

\begin{align}
  x_1' &= 2 \cdot \frac{x_1}{w} - 1 \\
  x_2' &= 2 \cdot \frac{x_2}{w} - 1 \\
  y_1' &= 2 \cdot \frac{y_1}{h} - 1 \\
  y_2' &= 2 \cdot \frac{y_2}{h} - 1
\end{align}

The normalized bounding box coordinates for subject and object are stacked to a vector of length 8. This vector is passed to a two-layer MLP which projects the coordinate vector to the desired embedding token. For the size of the hidden layer, we use half the embedding dimension.



\subsection{Binary Subject-Object Encoding}

Eq. 1 shows how subject and object location are encoded into the patch tokens before they are processed by DSFormer's transformer module. We use $r_{sbj}$ and $r_{obj}$ to represent the ratio of how much a specific patch is covered by the respective segmentation mask. Alternatively, we can replace the ratios with binary values that are 1 if the patch is covered partially by the segmentation mask and 0 otherwise. However, we did not observe significant performance improvements with this modification.

\subsection{Pretraining}

For a fair comparison, we did not pretrain DSFormer on precursor tasks. However, DSFormer can be pretrained on segmentation datasets by disabling the relation classifier and just classifying pairs of subject and object using the node classifier.

\section{Ablation Study}

This section contains additional results of our ablation study.

\Cref{tab:components} shows how the individual components of DSFormer benefit the overall score. In the table, a cross (\textcolor{red}{$\times$}) in the Masks column indicates that the actual segmentation mask was replaced by a rectangular segmentation box of the size of the related bounding box. If these rectangular masks are used, adding the semantic token results in a greater improvement than adding the location token. However, if the actual masks are used, the location token is more important. We assume that rectangular mask and location token encode the same information in two different ways and therefore adding the location token merely improves the model (+0.08\emph{mR@50}). On the other hand, actual segmentation mask and location token encode different information. Consequently, adding the location token drastically improves performance (+6.3\emph{mR@50}).

\setlength{\tabcolsep}{9pt}

\begin{table}[tb]
  \caption{Components. A cross (\textcolor{red}{$\times$}) in the Masks column means that instead of an actual segmentation mask, the enclosing bounding box region was used to create a rectangular segmentation mask. All experiments were run 3 times. The number behind the $\pm$ sign shows the standard deviation.}
  \label{tab:components}
  \centering
  \resizebox{\linewidth}{!}{
    \begin{tabular}{cccrr}
      \toprule
      Masks & Location Token & Semantic Token & mR@50 & mNgR@50 \\
      \midrule
      \textcolor{red}{$\times$} & \textcolor{red}{$\times$} & \textcolor{red}{$\times$} & 33.75 $\pm$ 0.67 & 54.47 $\pm$ 1.12 \\
      \textcolor{red}{$\times$} & \textcolor{red}{$\times$} & \textcolor{Green}{\checkmark} & 36.52 $\pm$ 0.55 & 56.34 $\pm$ 0.96 \\
      \textcolor{red}{$\times$} & \textcolor{Green}{\checkmark} & \textcolor{red}{$\times$} & 33.83 $\pm$ 1.15 & 53.97 $\pm$ 1.16 \\
      \textcolor{red}{$\times$} & \textcolor{Green}{\checkmark} & \textcolor{Green}{\checkmark} & 38.80 $\pm$ 0.79 & 60.77 $\pm$ 1.54 \\
      \textcolor{Green}{\checkmark} & \textcolor{red}{$\times$} & \textcolor{red}{$\times$} & 32.48 $\pm$ 0.33 & 51.80 $\pm$ 0.36 \\
      \textcolor{Green}{\checkmark} & \textcolor{red}{$\times$} & \textcolor{Green}{\checkmark} & 36.64 $\pm$ 0.47 & 58.36 $\pm$ 1.28 \\
      \textcolor{Green}{\checkmark} & \textcolor{Green}{\checkmark} & \textcolor{red}{$\times$} & 38.78 $\pm$ 0.70 & 61.75 $\pm$ 0.71 \\
      \textcolor{Green}{\checkmark} & \textcolor{Green}{\checkmark} & \textcolor{Green}{\checkmark} & \textbf{40.06 $\pm$ 0.56} & \textbf{64.05 $\pm$ 1.91} \\
      \bottomrule
      \end{tabular}
  }
\end{table}

\Cref{tab:nodeweight} shows the importance of our auxiliary node loss (Sec\onedot 3.6). Without it, the performance degrades. However, adding too much auxiliary loss also degrades performance.
\begin{table}[tb]
  \caption{Node loss weight. All experiments were run 3 times. The number behind the $\pm$ sign shows the standard deviation.}
  \label{tab:nodeweight}
  \centering
  \begin{tabular}{rrr}
    \toprule
    Node Loss Weight & mR@50 & mNgR@50 \\
    \midrule
    0.00 & 35.96 $\pm$ 2.36 & 61.04 $\pm$ 4.62 \\
    0.20 & \textbf{40.06 $\pm$ 0.56} & 64.05 $\pm$ 1.91 \\
    0.50 & 39.90 $\pm$ 0.76 & \textbf{64.19 $\pm$ 1.48} \\
    \bottomrule
    \end{tabular}
\end{table}




\Cref{tab:embdim} shows that with increasing embedding dimension, performance is improved. However, the improvements begin to converge with higher embedding dimension sizes.

\begin{table}[H]
  \caption{Embedding dimension size for the transformer module. All experiments were run 3 times. The number behind the $\pm$ sign shows the standard deviation.}
  \label{tab:embdim}
  \centering
  \begin{tabular}{rrr}
    \toprule
    Embedding Dimension & mR@50 & mNgR@50 \\
    \midrule
    192 & 38.01 $\pm$ 1.27 & 60.67 $\pm$ 1.47 \\
    256 & 39.24 $\pm$ 1.00 & 62.88 $\pm$ 0.31 \\
    384 & \textbf{40.06 $\pm$ 0.56} & \textbf{64.05 $\pm$ 1.91} \\
    \bottomrule
    \end{tabular}
\end{table}




\section{Example Images}

The images below show examle outputs from DSFormer. The blue arrows are the ground truth annotations.
For every relation, DSFormer assigns a score to each predicate and the predicates can be ranked within a relation.
In the shown example images, the ranks are shown as numbers behind the predicate names.
A low number means that DSFormer estimates the predicate to be more suitable than other predicates with a higher rank.
The highest possible number is 56 (the total number of predicates in the PSG dataset).
The text in green is the ground truth predicate label for the relation.

\def\imgSpace{\vspace{18pt}}

\begin{figure}[H]
  \centering
  \begin{subfigure}[t]{0.45\linewidth}
    \includegraphics[width=1.0\linewidth]{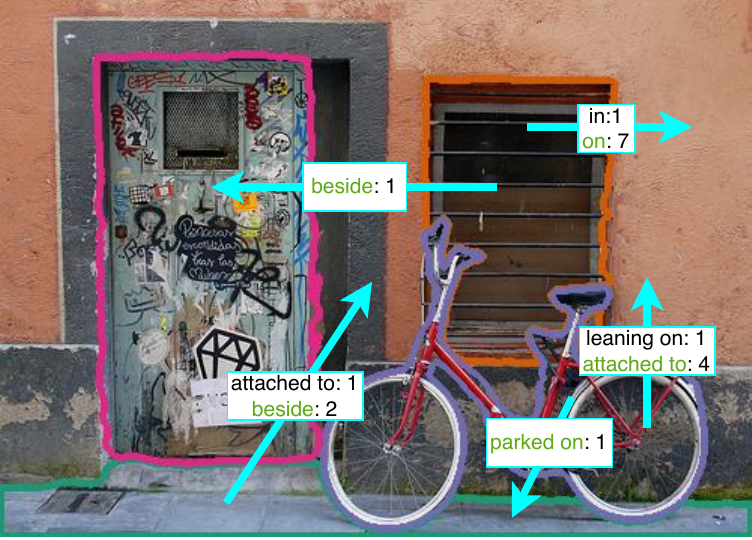}
    \centering
  \end{subfigure}
  \hfill
  \begin{subfigure}[t]{0.45\linewidth}
    \includegraphics[width=1.0\linewidth]{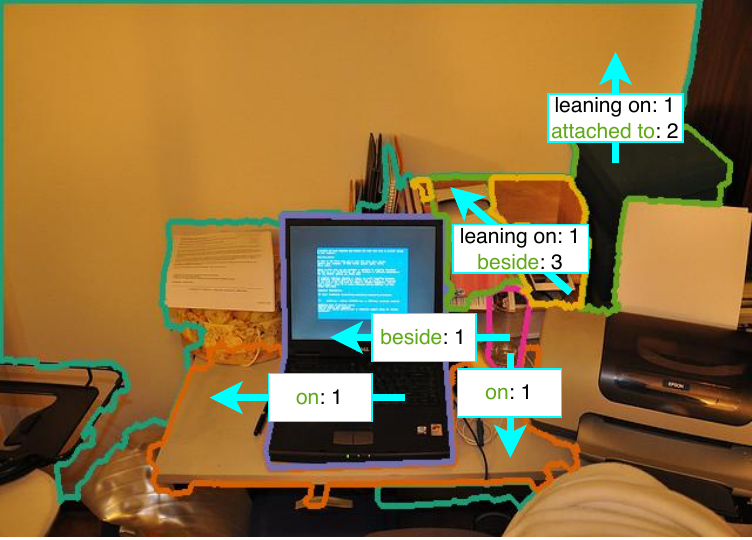}
    \centering
  \end{subfigure}
\end{figure}

\vspace{-30pt}

\begin{figure}[H]
  \begin{subfigure}[t]{0.45\linewidth}
    \includegraphics[width=1.0\linewidth]{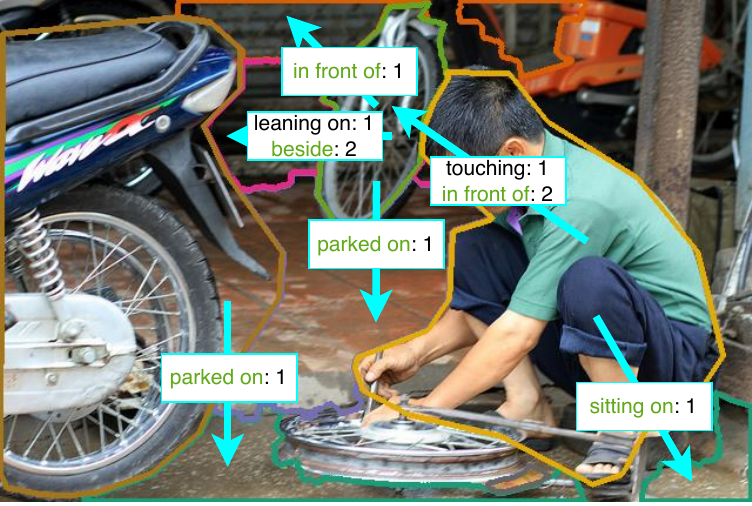}
    \centering
  \end{subfigure}
  \hfill
  \begin{subfigure}[t]{0.45\linewidth}
    \includegraphics[width=1.0\linewidth]{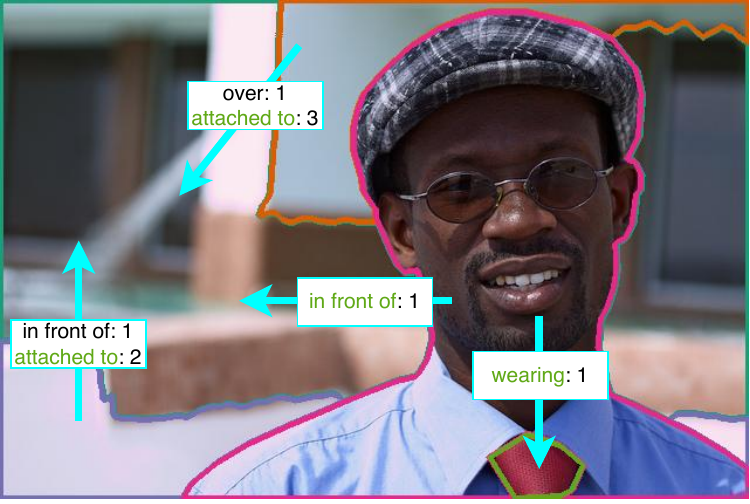}
    \centering
  \end{subfigure}
\end{figure}

\vspace{-30pt}

\begin{figure}[H]
  \begin{subfigure}[t]{0.45\linewidth}
    \includegraphics[width=1.0\linewidth]{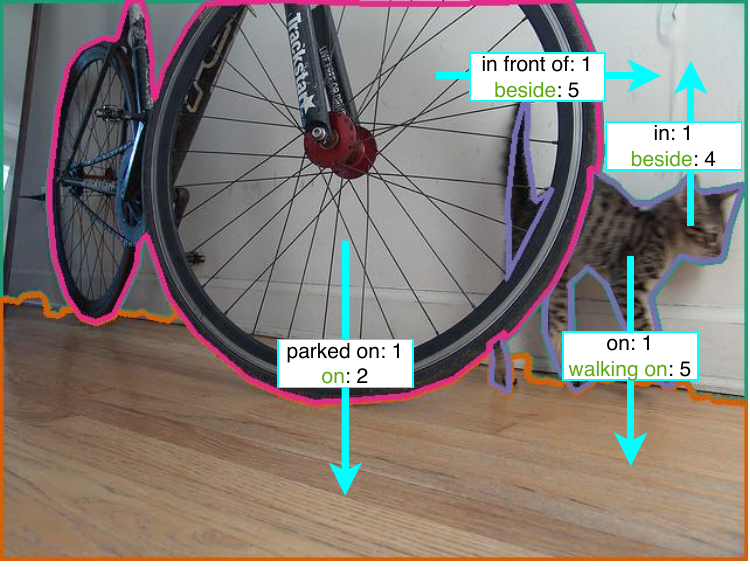}
    \centering
  \end{subfigure}
  \hfill
  \begin{subfigure}[t]{0.45\linewidth}
    \includegraphics[width=1.0\linewidth]{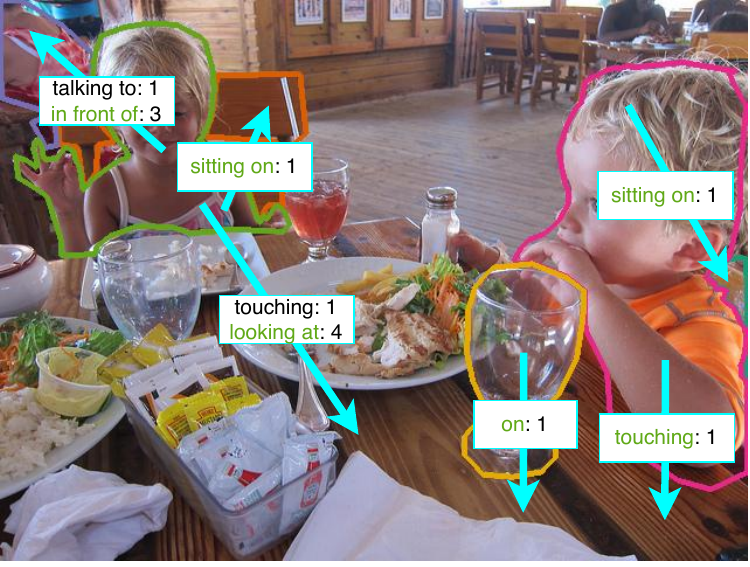}
    \centering
  \end{subfigure}
\end{figure}

\vspace{-30pt}

\begin{figure}[H]
  \begin{subfigure}[t]{0.45\linewidth}
    \includegraphics[width=1.0\linewidth]{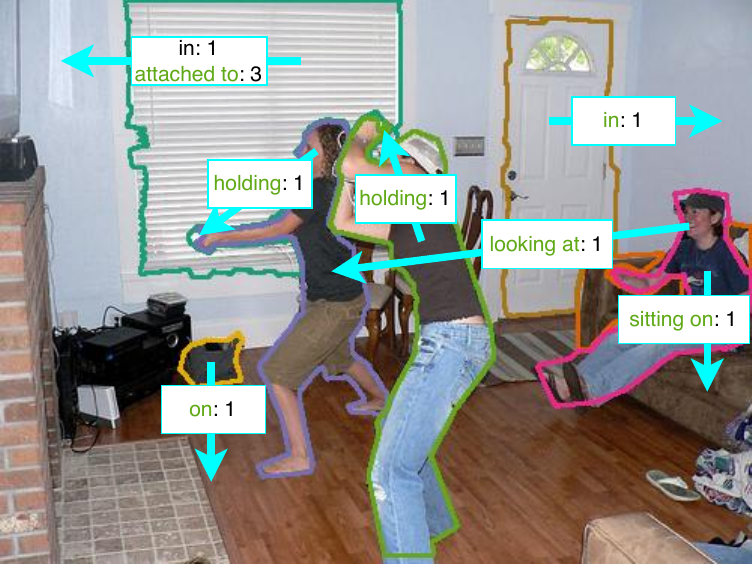}
    \centering
  \end{subfigure}
  \hfill
  \begin{subfigure}[t]{0.45\linewidth}
    \includegraphics[width=1.0\linewidth]{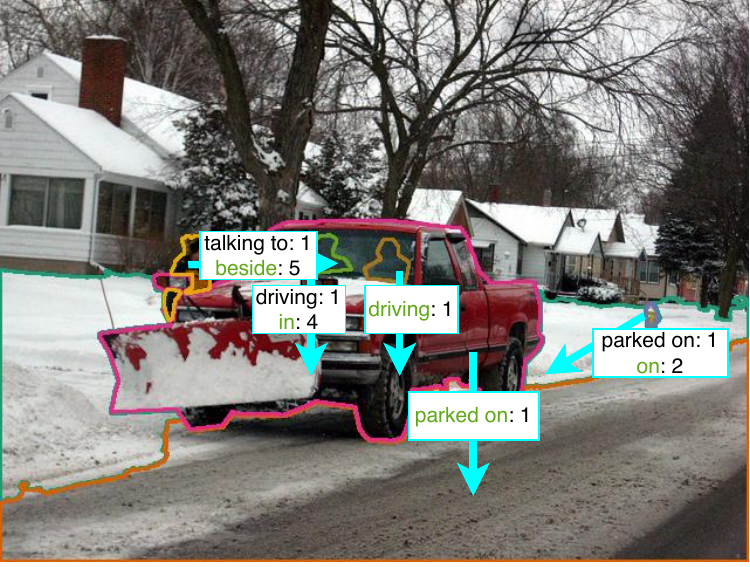}
    \centering
  \end{subfigure}
\end{figure}


\begin{thebibliography}{10}
\providecommand{\url}[1]{\texttt{#1}}
\providecommand{\urlprefix}{URL }
\providecommand{\doi}[1]{https://doi.org/#1}

\bibitem{detr}
Carion, N., Massa, F., Synnaeve, G., Usunier, N., Kirillov, A., Zagoruyko, S.: End-to-end object detection with transformers. In: Vedaldi, A., Bischof, H., Brox, T., Frahm, J.M. (eds.) Computer Vision -- ECCV 2020. pp. 213--229. Springer International Publishing, Cham (2020)

\bibitem{relwork_multiproto}
Chen, L., Song, Y., Cai, Y., Lu, J., Li, Y., Xie, Y., Wang, C., He, G.: Multi-prototype space learning for commonsense-based scene graph generation. Proceedings of the AAAI Conference on Artificial Intelligence  \textbf{38}(2),  1129--1137 (Mar 2024). \doi{10.1609/aaai.v38i2.27874}, \url{https://ojs.aaai.org/index.php/AAAI/article/view/27874}

\bibitem{kern}
Chen, T., Yu, W., Chen, R., Lin, L.: Knowledge-embedded routing network for scene graph generation. In: 2019 IEEE/CVF Conference on Computer Vision and Pattern Recognition (CVPR). pp. 6156--6164 (2019). \doi{10.1109/CVPR.2019.00632}

\bibitem{mask2former}
Cheng, B., Misra, I., Schwing, A.G., Kirillov, A., Girdhar, R.: Masked-attention mask transformer for universal image segmentation. arXiv  (2021)

\bibitem{vit}
Dosovitskiy, A., Beyer, L., Kolesnikov, A., Weissenborn, D., Zhai, X., Unterthiner, T., Dehghani, M., Minderer, M., Heigold, G., Gelly, S., Uszkoreit, J., Houlsby, N.: An image is worth 16x16 words: Transformers for image recognition at scale. In: International Conference on Learning Representations (2021), \url{https://openreview.net/forum?id=YicbFdNTTy}

\bibitem{maskrcnn}
He, K., Gkioxari, G., Dollár, P., Girshick, R.: {Mask R-CNN}. In: 2017 IEEE International Conference on Computer Vision (ICCV). pp. 2980--2988 (2017). \doi{10.1109/ICCV.2017.322}

\bibitem{oneformer}
Jain, J., Li, J., Chiu, M., Hassani, A., Orlov, N., Shi, H.: Oneformer: One transformer to rule universal image segmentation. In: 2023 IEEE/CVF Conference on Computer Vision and Pattern Recognition (CVPR). pp. 2989--2998 (2023). \doi{10.1109/CVPR52729.2023.00292}

\bibitem{hotr}
Kim, B., Lee, J., Kang, J., Kim, E.S., Kim, H.J.: {HOTR}: End-to-end human-object interaction detection with transformers. In: CVPR. IEEE (2021)

\bibitem{panseg}
Kirillov, A., He, K., Girshick, R., Rother, C., Dollár, P.: Panoptic segmentation. In: 2019 IEEE/CVF Conference on Computer Vision and Pattern Recognition (CVPR). pp. 9396--9405 (2019). \doi{10.1109/CVPR.2019.00963}

\bibitem{visual_genome}
Krishna, R., Zhu, Y., Groth, O., Johnson, J., Hata, K., Kravitz, J., Chen, S., Kalantidis, Y., Li, L.J., Shamma, D.A., Bernstein, M.S., Fei-Fei, L.: Visual genome: Connecting language and vision using crowdsourced dense image annotations. International Journal of Computer Vision  \textbf{123}(1),  32--73 (2017). \doi{10.1007/s11263-016-0981-7}, \url{https://doi.org/10.1007/s11263-016-0981-7}

\bibitem{maskdino}
Li, F., Zhang, H., Xu, H., Liu, S., Zhang, L., Ni, L.M., Shum, H.Y.: {Mask DINO}: Towards a unified transformer-based framework for object detection and segmentation. In: Proceedings of the IEEE/CVF Conference on Computer Vision and Pattern Recognition (CVPR). pp. 3041--3050 (June 2023)

\bibitem{relwork_semproto}
Li, L., Ji, W., Wu, Y., Li, M., Qin, Y., Wei, L., Zimmermann, R.: Panoptic scene graph generation with semantics-prototype learning. Proceedings of the AAAI Conference on Artificial Intelligence  \textbf{38}(4),  3145--3153 (Mar 2024). \doi{10.1609/aaai.v38i4.28098}, \url{https://ojs.aaai.org/index.php/AAAI/article/view/28098}

\bibitem{relwork_invarlearn}
Li, L., Qin, Y., Ji, W., Zhou, Y., Zimmermann, R.: Domain-wise invariant learning for panoptic scene graph generation. In: ICASSP 2024 - 2024 IEEE International Conference on Acoustics, Speech and Signal Processing (ICASSP). pp. 3165--3169 (2024). \doi{10.1109/ICASSP48485.2024.10447193}

\bibitem{relwork_cktrcm}
Liang, N., Liu, Y., Sun, W., Xia, Y., Wang, F.: Ckt-rcm: Clip-based knowledge transfer and relational context mining for unbiased panoptic scene graph generation. In: ICASSP 2024 - 2024 IEEE International Conference on Acoustics, Speech and Signal Processing (ICASSP). pp. 3570--3574 (2024). \doi{10.1109/ICASSP48485.2024.10446810}

\bibitem{fpn}
Lin, T.Y., Dollár, P., Girshick, R., He, K., Hariharan, B., Belongie, S.: Feature pyramid networks for object detection. In: 2017 IEEE Conference on Computer Vision and Pattern Recognition (CVPR). pp. 936--944 (2017). \doi{10.1109/CVPR.2017.106}

\bibitem{coco}
Lin, T.Y., Maire, M., Belongie, S., Hays, J., Perona, P., Ramanan, D., Doll{\'a}r, P., Zitnick, C.L.: {Microsoft COCO}: Common objects in context. In: Fleet, D., Pajdla, T., Schiele, B., Tuytelaars, T. (eds.) Computer Vision -- ECCV 2014. pp. 740--755. Springer International Publishing, Cham (2014)

\bibitem{gpsnet}
Lin, X., Ding, C., Zeng, J., Tao, D.: Gps-net: Graph property sensing network for scene graph generation. In: Proceedings of the IEEE/CVF Conference on Computer Vision and Pattern Recognition (CVPR) (June 2020)

\bibitem{haystack}
Lorenz, J., Barthel, F., Kienzle, D., Lienhart, R.: Haystack: A panoptic scene graph dataset to evaluate rare predicate classes. In: Proceedings of the IEEE/CVF International Conference on Computer Vision (ICCV) Workshops. pp. 62--70 (October 2023)

\bibitem{first_scenegraph}
Lu, C., Krishna, R., Bernstein, M., Fei-Fei, L.: Visual relationship detection with language priors. In: Leibe, B., Matas, J., Sebe, N., Welling, M. (eds.) Computer Vision – {ECCV} 2016. pp. 852--869. Lecture Notes in Computer Science, Springer International Publishing (2016). \doi{10.1007/978-3-319-46448-0_51}

\bibitem{arbitrary_keypoints}
Ludwig, K., Harzig, P., Lienhart, R.: Detecting arbitrary intermediate keypoints for human pose estimation with vision transformers. In: 2022 IEEE/CVF Winter Conference on Applications of Computer Vision Workshops (WACVW). pp. 663--671 (2022). \doi{10.1109/WACVW54805.2022.00073}

\bibitem{fasterrcnn}
Ren, S., He, K., Girshick, R., Sun, J.: {Faster R-CNN}: Towards real-time object detection with region proposal networks. IEEE Transactions on Pattern Analysis and Machine Intelligence  \textbf{39}(6),  1137--1149 (2017). \doi{10.1109/TPAMI.2016.2577031}

\bibitem{objects365}
Shao, S., Li, Z., Zhang, T., Peng, C., Yu, G., Zhang, X., Li, J., Sun, J.: Objects365: A large-scale, high-quality dataset for object detection. In: 2019 IEEE/CVF International Conference on Computer Vision (ICCV). pp. 8429--8438 (2019). \doi{10.1109/ICCV.2019.00852}

\bibitem{vctree}
Tang, K., Zhang, H., Wu, B., Luo, W., Liu, W.: Learning to compose dynamic tree structures for visual contexts. In: Conference on Computer Vision and Pattern Recognition (2019)

\bibitem{pairnet}
Wang, J., Wen, Z., Li, X., Guo, Z., Yang, J., Liu, Z.: Pair then relation: {Pair-Net} for panoptic scene graph generation. \doi{10.48550/arXiv.2307.08699}, \url{http://arxiv.org/abs/2307.08699}

\bibitem{as1b}
Wang, W., Shi, M., Li, Q., Wang, W., Huang, Z., Xing, L., Chen, Z., Li, H., Zhu, X., Cao, Z., et~al.: The all-seeing project: Towards panoptic visual recognition and understanding of the open world. arXiv preprint arXiv:2308.01907  (2023)

\bibitem{imp}
Xu, D., Zhu, Y., Choy, C., Fei-Fei, L.: Scene graph generation by iterative message passing. In: Computer Vision and Pattern Recognition (CVPR) (2017)

\bibitem{psg}
Yang, J., Ang, Y.Z., Guo, Z., Zhou, K., Zhang, W., Liu, Z.: Panoptic scene graph generation. In: ECCV (2022)

\bibitem{relwork_4dpsgg}
Yang, J., CEN, J., PENG, W., Liu, S., Hong, F., Li, X., Zhou, K., Chen, Q., Liu, Z.: 4d panoptic scene graph generation. In: Oh, A., Naumann, T., Globerson, A., Saenko, K., Hardt, M., Levine, S. (eds.) Advances in Neural Information Processing Systems. vol.~36, pp. 69692--69705. Curran Associates, Inc. (2023), \url{https://proceedings.neurips.cc/paper_files/paper/2023/file/dc6319dde4fb182b22fb902da9418566-Paper-Conference.pdf}

\bibitem{motifs}
Zellers, R., Yatskar, M., Thomson, S., Choi, Y.: Neural motifs: Scene graph parsing with global context. In: Conference on Computer Vision and Pattern Recognition (2018)

\bibitem{ietrans}
Zhang, A., Yao, Y., Chen, Q., Ji, W., Liu, Z., Sun, M., Chua, T.S.: Fine-grained scene graph generation with data transfer. In: Avidan, S., Brostow, G., Ciss{\'e}, M., Farinella, G.M., Hassner, T. (eds.) Computer Vision -- ECCV 2022. pp. 409--424. Springer Nature Switzerland, Cham (2022)

\bibitem{hilo}
Zhou, Z., Shi, M., Caesar, H.: {HiLo}: Exploiting high low frequency relations for unbiased panoptic scene graph generation. In: Proceedings of the IEEE/CVF International Conference on Computer Vision (ICCV). pp. 21637--21648 (October 2023)

\bibitem{vlprompt}
Zhou, Z., Shi, M., Caesar, H.: Vlprompt: Vision-language prompting for panoptic scene graph generation. arXiv preprint arXiv:2311.16492  (2023)

\end{thebibliography}


\begin{thebibliography}{1}
\providecommand{\url}[1]{\texttt{#1}}
\providecommand{\urlprefix}{URL }
\providecommand{\doi}[1]{https://doi.org/#1}

\bibitem{adamw}
Loshchilov, I., Hutter, F.: Decoupled weight decay regularization (2019)

\bibitem{transformer}
Vaswani, A., Shazeer, N., Parmar, N., Uszkoreit, J., Jones, L., Gomez, A.N., Kaiser, L., Polosukhin, I.: Attention is all you need. In: Guyon, I., Luxburg, U.V., Bengio, S., Wallach, H., Fergus, R., Vishwanathan, S., Garnett, R. (eds.) Advances in Neural Information Processing Systems. vol.~30. Curran Associates, Inc. (2017)

\end{thebibliography}
\putbib[supp]
\end{bibunit}

\end{document}